\documentclass[letterpaper]{article}
\usepackage{aaai}
\usepackage{times}
\usepackage{helvet}
\usepackage{courier}
\usepackage{latexsym}
\usepackage{multirow}
\usepackage{CJKutf8}
\newcommand{\chinese}[1]{\begin{CJK*}{UTF8}{gkai}#1\end{CJK*}}
\renewcommand{\paragraph}[1]{\noindent\textbf{#1}}
\usepackage{enumitem}
\setlist[itemize]{topsep=0pt,itemsep=0pt,parsep=0pt,before=\vspace{3pt}} 
\setlist[enumerate]{topsep=0pt,itemsep=0pt,parsep=0pt,before=\vspace{3pt}} 
\usepackage{graphicx}
\usepackage{array}
\newcolumntype{L}[1]{>{\raggedright\let\newline\\\arraybackslash}m{#1}}
\newcolumntype{C}[1]{>{\centering\let\newline\\\arraybackslash}m{#1}}
\newcolumntype{R}[1]{>{\raggedleft\let\newline\\\arraybackslash}m{#1}}
\usepackage{microtype}
\usepackage{todonotes}
\usepackage{xcolor}
\usepackage[utf8]{inputenc}

\begin{document}
%
\title{Stylistic Retrieval-based Dialogue System with Unparallel Training Data}
\author{Hao Fu,\textsuperscript{\rm 1}
Yan Wang,\textsuperscript{\rm 2}
Ruihua Song,\textsuperscript{\rm 3}
Tianran Hu,\textsuperscript{\rm 4}
Jian-Yun Nie\textsuperscript{\rm 5}\\
\textsuperscript{\rm 1}Meituan Group \\
\textsuperscript{\rm 2}Tencent AI Lab \\
\textsuperscript{\rm 3}Gaoling School of Artificial Intelliegent, Renmin University of China \\
\textsuperscript{\rm 4}Computer Science, College of William \& Mary \\
\textsuperscript{\rm 5}Department of Computer Science and Operations Research, Universite De Montreal \\
fuch@mail.ustc.edu.cn, brandenwang@tencent.com,  rsong@ruc.edu.cn, thu03@wm.edu, nie@iro.umontreal.ca}
\maketitle
\begin{abstract}
\begin{quote}
The ability of a dialog system to express consistent language style during conversations
has a direct, positive impact on its usability
and on user satisfaction. Although previous studies have demonstrated that style transfer is feasible with a large amount of parallel data, it is often impossible to collect such data for different styles. In this paper, instead of manually constructing conversation data with a certain style, we propose a flexible framework that adapts a generic retrieval-based dialogue system to mimic the language style of a specified persona without any parallel data. Our approach is based on automatic generation of stylized data by learning the usage of jargon, and then rewriting the generic conversations to a stylized one by incorporating the jargon. In experiments we implemented dialogue systems with five distinct language styles, and the result shows our framework significantly outperforms baselines in terms of the average score of responses' relevance and style degree, and content diversity. A/B testing on a commercial chatbot shows that users are more satisfied with our system. This study demonstrates the feasibility of building stylistic dialogue systems by simple data augmentation.
\end{quote}
\end{abstract}

\section{Introduction}
\label{sec:intro}

\begin{figure}
\centering
\includegraphics[width=3in]{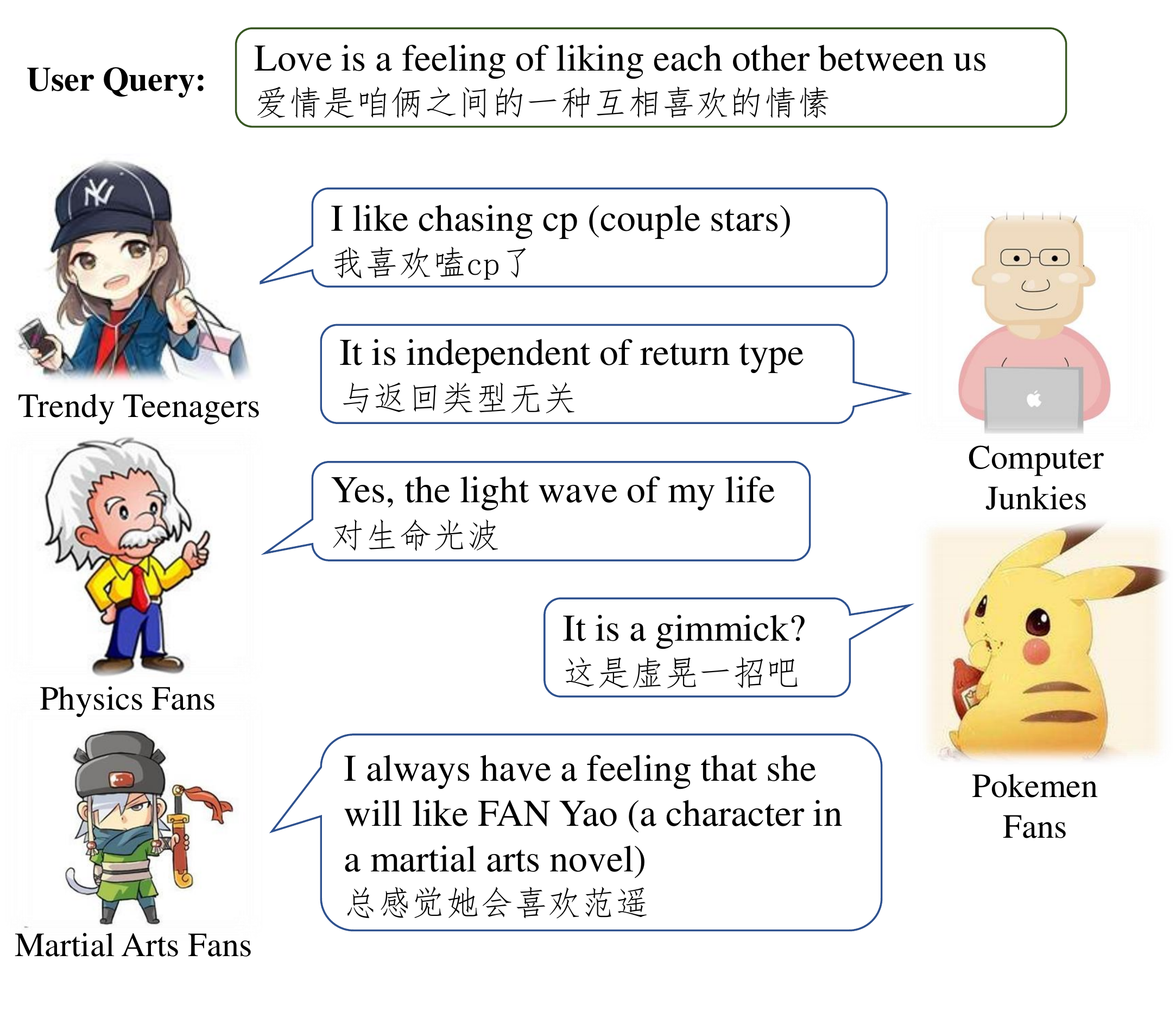}
\vspace{-18pt}
\caption{Responses of our chitchat dialogue systems with different character settings.}
\label{fig:application}
\vspace{-6pt}
\end{figure}

\noindent Many researches show that incorporating role-specific language style makes a dialogue system more human-like and attractive~\shortcite{li2016persona}.  As shown in Figure~\ref{fig:application}, in stylistic dialogue generation, the responses are expected to be not only relevant to the context, but also consistent with the desired style. Recently, on the basis of parallel data of an utterance and its stylized version, some studies show the effect of stylistic dialogue generation to account for emotions \shortcite{DBLP:conf/aaai/ZhouHZZL18}, response attitudes \shortcite{niu2018polite}, personas \cite{li2016persona}, and gender identification. \shortcite{IGRL,su2021prototype}.

Despite their success, previous studies are generally conducted in a fully supervised setting requiring stylistic dialogue corpus.
However, it is expensive and not always feasible to create such a corpus, due to the lack of parallel data which consists of many triplets in \{style, context, response\} format. For a target style, previous work either looks for specific data sources such as movie transcripts~\shortcite{jena2017enterprise}, or recruits human crowd-workers to create corpora~\shortcite{zhang2018personalizing}. On the other hand, although most state-of-the-art business chatbots are built upon information retrieval technology, there isn't any practical way to train a stylistic retrieval-based dialogue agent. The only way to prevent the dialogue agent from contradicting its desired style is to carefully annotate all the corpus and deliberately filter out those contradicted pairs, which is expensive and inefficient.

In this paper, we present a light-weight approach 
that enables a \textit{generic} retrieval-based dialogue system (which does not embed any persona) to mimic the language style of a specific role.
Our framework only requires minimum human inputs (e.g., 500 keywords) for each style, and does not need any paralleled data. Instead of building an entirely new dialogue system for each style from scratch, our framework functions as a ``patch'' to stylize a generic retrieval-based dialogue system. The main idea is to rewrite a generic dialogue repository into a stylistic one. Specifically, we consider that a target style is represented by a set of jargon phrases (words that are frequently used by a particular group), and propose a jargon-based rewriting schema to tint the responses in dialogue repository with the desired style. 

One problem in this rewriting schema is that the jargon in the rewritten stylistic responses may have never occurred in the original repository, and the re-ranking model trained on a generic repository may not understand them. To alleviate this problem, we further propose an alignment process to align each stylized response to an internal response that substitute the jargon phrases with synonyms in plain language. In the re-ranking process the model computes matching degrees based on these internal responses and then the aligned stylized responses are proposed as the final output. 

To evaluate our framework, we implement five dialogue agents with distinct personas and styles -- trendy teenagers who prefer informal expression and slang, computer junkies who adopt many technical terms in their language, physics fans who frequently use physics jargon, Pok\'emon fans who like using game jargon, and martial arts fans who frequently mention tricks or characters in martial arts novels. The stylistic dialogues are evaluated by both human annotators and automatic metrics. Extensive experiments show that our system properly balances the trade-off between relevance and style intensity, and yields responses with better comprehensive performance.
Online A/B test indicates that our implemented chatbot with the trendy teenager character setting improves users engagement.

This paper has three main contributions:
\begin{enumerate}
    \item We endow the capability of responding in specific language style to retrieval-based dialogue systems. More importantly, this method can be used to patch any generic dialogue systems for various personas, without any extra fine-tuning or re-training of dialogue models.  
    \item Our method gets rid of the dependency of paralleled data. Instead we only need a small set of jargon phrases and synonyms to rewrite the dialogue repository. 
    \item Offline experimental results and online A/B test show the superiority of our approach, as well as the effect of stylistic response on user engagement. 
\end{enumerate}

In the remaining part, we review related work. Then we describe our approach in details. 
Experimental results of both offline evaluation and online A/B test are presented finally.

\begin{figure*}
\centering
\includegraphics[width=0.95\textwidth]{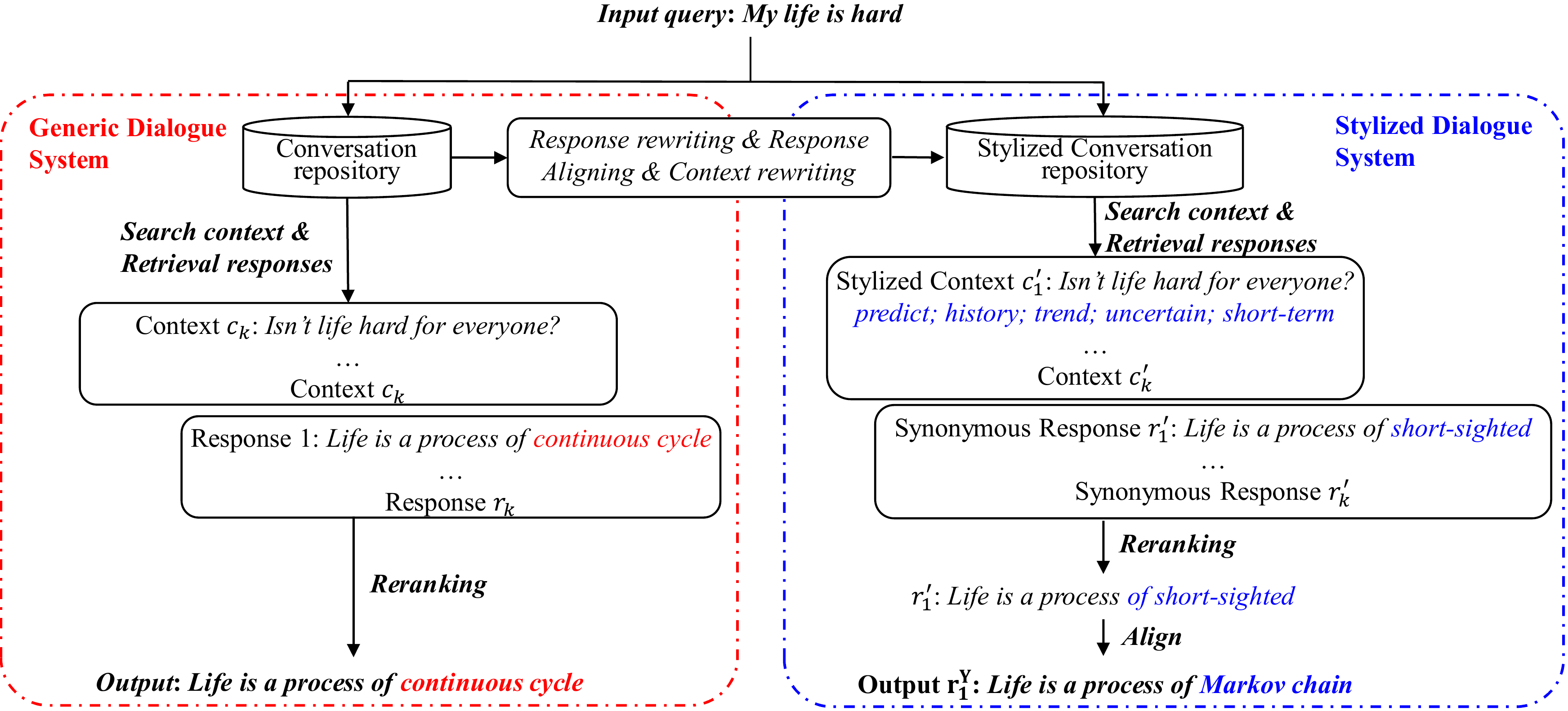}
\small
\vspace{-6pt}
\caption{Framework of a generic dialogue system and our stylized dialogue system.}
\label{fig:framework}
\end{figure*}

\section{Related Work}
\label{sec:related-work}

\subsection{Retrieval-based Dialogue Systems}
Recently there has been growing interest in building intelligent chatbots to complete tasks or chitchat. This paper focuses on retrieval-based dialogue system that first recall several response candidates from dialogue corpus via lexical-similarity between context, and then learns a semantic matching function between the conversation context and response candidates for response selection~\shortcite{yan2016learning,wu2017sequential,zhou2018multi,zhang2018modeling,tao2019multi}
Such systems generally produce more fluent and diverse responses and are 
widely used in commercial chatbots~\cite{shum2018eliza}.

\subsection{Persona-based Dialogue System}
\citeyear{li2016persona} introduce persona-based dialogue models which capture speaker's role and represent it as distributed embeddings. 
Approaches have been proposed to learn a persona model from user profiles~\shortcite{zhang2018personalizing,luo2019learning} and personalized dialogue~\shortcite{herzig2017neural,zhang2019neural,madotto2019personalizing,song2020generate,song2021bob}. However, most persona-based models focus on the consistency of dialogue with the persona, but do not address the language style problem, which is an important aspect of persona-based dialogue..

Some recent studies focus on specific language styles. \citeyear{jena2017enterprise} introduce a Star Trek chatbot that imitates the speaking style of characters in the show. The chatbot is composed of two models that handle Star Trek style input and everyday conversation separately. \citeyear{niu2018polite} propose a set of models for generating polite dialogues. The models are guided by a politeness classifier in generating responses. 
\citeyear{su2021prototype} first extract a non-stylistic prototype from a generic dialogue system and then generate stylistic response via GPT2. \citeyear{zheng2020stylized} and \citeyear{zhuneural} attempt to using monolingual stylistic data to increase the style intensity of dialogue response. 
We notice that all these studies focus on a single style, and are difficult to extend to other styles. Compared to these methods, our framework is retrieval-based, and can be used cover for a broader range of personas due to its light-weight nature.

\subsection{Text Style Transfer}
Style transfer of text is a related task of stylizing dialogue systems. The task is transferring a sentence to a specified language style, while keeping the content of the sentence unchanged. A representative approach is learning disentangled latent representations of content and style. \citeyear{shen2017style} propose a cross-aligned auto-encoder with shared content and separated style representations. \citeyear{hu2017toward} adversarially combine variational auto-encoder and style classifier. Several studies leverage adversarial networks with different discriminators and training objectives~\shortcite{fu2018style,yang2018unsupervised,john2019disentangled,liu2021learning,goyal2021multi}.

Another approach separates content and style via explicit sequence editing. \citeyear{fu2019rethinking} show that style transfer can be largely achieved by lexical substitution. \citeyear{li2018delete} propose a ``delete, retrieve, and generate'' framework that replaces style-related words and re-generate the sentence for better fluency. Similar approaches based on pre-trained language model or reinforcement learning are also explored~\shortcite{xu2018unpaired,sudhakar2019transforming,wu2019hierarchical,krishna2020reformulating,li2020unsupervised}.

Our approach to stylize dialogue is consistent with the above studies (e.g. with the observation of \citeyear{fu2019rethinking} on lexical substitution). However, we do not employ a heavy mechanism to implement the transfer but exploit a small set of style-related terms to generate stylized sentences. We compare with  \cite{shen2017style} and \cite{li2018delete} in experiments.
\begin{table}
\centering
\begin{tabular}{p{0.8in}p{0.98in}p{1in}}
Style & Jargon phrase & Synonym \\
\hline
\multirow{2}{*}{Trendy Teenager} & \chinese{极好的} (snatched) & really good \\
& \chinese{智商税} (stupid tax) & pay for ignorance \\
\hline
\multirow{2}{*}{Computer} & access token & key  \\
& transfer learning & learn by analogy  \\
\hline
\multirow{2}{*}{Physics} & polarization & single direction \\
 & sonic boom & very fast\\
\hline
\multirow{2}{*}{Pok\'emon} & fire blast & burn  \\
 & Squirtle & turtle \\
\hline
\multirow{2}{*}{Martial Arts} & \chinese{裘千尺} \footnotemark & an antagonist  \\
 & (Qiu Chianchi) &  \\
\end{tabular}
\caption{Samples of annotated parallel phrases.}
\label{tab:sample-align-phrase}
\vspace{-12pt}
\end{table}

\section{Our Proposed Framework}

\subsection{Main Idea}
Our proposed framework is shown in Figure~\ref{fig:framework}. To construct a stylized retrieval-based dialogue system, our main idea is to rewrite the source corpus $\mathcal{X}$ to a stylized corpus $X^Y$. The rewriting is based on a jargon - synonym set that contains about 500 jargon phrases and their synonyms in plain language (annotated by human). The first step is \textbf{Response Rewriting} that transforms every response $r$ in $\mathcal{X}$ to a stylized response $r^Y$ with jargon phrases via a style transfer model. After response rewriting, two main challenges still remain: 1) The ranking model trained on a generic repository may not understand the stylized responses $r^Y$, and 2) the stylized responses $r^Y$ may be irrelevant to the original context $c$. To alleviate these problems, we further propose two modules, \textbf{Response Alignment} module and \textbf{Context Rewriting} module, respectively. The former attempts to align $r^Y$ to a synonymous response $r'$ in plain language through the jargon - synonym set, so that the reranking are based on $r'$ instead of the $r^Y$. For the latter, we rewrite the context of $c$ to $c'$ by adding some context words that are related to the jargon so that $c'$ is more relevant to the stylized response. For example, for the jargon ``Markov chain'', we rewrite its origin context \textit{``Isn't life hard for everyone?''} to \textit{``Isn't life hard for everyone? predict; history; trend; uncertain; short-term''}.

After Rewriting and Alignment, we can “patch” a generic retrieval-based dialogue system to adapt new language styles. As shown in Figure~\ref{fig:framework}, after replacing the all context-response pairs $(c, r)$ in a generic conversation repository $\mathcal{X}$ to stylized pairs $(c', r')$,
the dialogue system searches for stylized contexts $c'$ and selects a synonymous response $r_{i}^{'}$ as an intermediate output. Then, the corresponded stylized response $r_i^Y$ is used as the response of input conversation.

\subsection{Jargon phrases \& synonyms} 
\label{sec:Jargon phrases}
We collect a set of jargon phrases from topic-specific websites and ask human experts to annotate them with synonyms in plain language. For instance, for the trendy teenager persona, we collected a list of Internet slang words from Fanjian\footnote{http://www.fanjian.net/}, a website maintaining a large database of Chinese Internet slang. We took a slang word as a jargon phrase $s^Y$. For each slang word, three human annotators were asked to assign a synonymous phrase $s^X$ in plain language. The annotators reviewed and discussed each other's annotation until they reached an agreement. Insulting slang words were removed. Samples of the phrases are presented in Table \ref{tab:sample-align-phrase}. The synonym phrases mostly explain certain aspects of the jargon phrases in plain language. We collected on average 523 parallel jargon phrase - synonym pairs per style.

\section{Approach}
In this section, we introduce detailed procedures of \textbf{Response Rewriting}, \textbf{Response Alignment}, and \textbf{Context Rewriting}, given a set of jargon phrases and synonyms.
The specified language style is implemented in the first step. The second step guarantees the synonymity of $r'$and $r^Y$. 
The third one fixes the relevance between $c'$ and $r'$. We further describe rewriting response and rewriting context in details.

\subsection{Response Rewriting}
\label{sec:main-corpus-response}
In this section, we introduce detailed procedures of rewriting responses by jargon phrases. Given an original response $r$, the goal of rewriting is to obtain a fluent and stylized response $r^Y$. A trivial way to achieve the goal is letting $r^Y$ be a common sentence like ``I am multi-threaded'' for any $r$. To avoid this, we require that $r^Y$ has as many overlapping words with $r$ as possible so that it is more specific. We generate a number of candidates for $r^Y$, and then filter out ungrammatical ones by a fine-tuned BERT model. 

\subsubsection{Generating Candidate Response}
\label{sec:main-corpus-response-candidate}

Given a response $r = \{w_1, w_2, ..., w_n\}$, which is a sequence of words, we select a sub-sequence $s_{i,j} = \{w_i, w_{i+1}, ..., w_j\}$ and substitute it with a jargon phrase $s^Y$. The resulting response $r^Y = \{w_1, ..., w_{i-1}, s^Y, w_{j+1}, ..., w_n\}$ is proposed as a candidate for further filtering.

This process is similar to that used in existing style transfer models~\shortcite{li2018delete,sudhakar2019transforming,wu2019mask}. These models substitute style-related words and keep others, so that the content is unchanged. To increase the coverage, we allow arbitrary words to be substituted. This relaxation yields $\mathcal{O}(n^2)$ possible substitutions, so we adopt the following constraints for pruning:

\begin{itemize}
    \item \paragraph{Word constraint:}
We only consider words and numbers for substitution. Punctuation marks are not substituted, because they play an important role in grammar. This constraint avoids generating grammatical errors by the change of punctuation.
    \item \paragraph{Length constraint:}
We try to keep $s_{i,j}$ short, so that most words of $r$ are preserved. This choice is made in order to avoid altering too much the initial sentence. Specifically, we limit the length of $s_{i,j}$ to be less than 5 words.
    \item \paragraph{Context constraint:}
We also require the sub-sequence $s_{i,j}$ to share the same preceding or succeeding words with a jargon phrase. For example, suppose we find sentences like ``... is fully multi-threaded ...'' in the stylized corpus $\mathcal{T}$. We may consider applying the substitution only if $s_{i,j}$ succeeds the words ``is fully''. 
This constraint helps filter out most of obviously ungrammatical candidates.
\end{itemize}

\subsubsection{Response Filtering}
\label{sec:main-corpus-response-filter}
We further filter stylized response candidates that are not fluent via a language model. Since a stylized response candidate is not an entirely new sentence, we only need to estimate the fluency of the substituted part.

We start with the pre-trained BERT model~\shortcite{devlin2019BERT}, and fine-tune it with a non-conversational stylized corpus $\mathcal{T}$. We mask the jargon phrases in $\mathcal{T}$ and train the model to recover them. Then we use this model to predict the probability of $s^Y$ in a given $r^Y$: $p(s^Y|r^Y)$. The candidate responses with low $p(s^Y|r^Y)$ will be filtered.  

\subsection{Aligning Responses}
After the \textbf{Response Rewriting} process, it is possible that the generated stylized response $r^Y$ is not synonymous to the original response $r$.
Moreover, the re-ranking model in the generic dialogue system may not understand $r^Y$ as well.
To alleviate these problems, we substitute stylized phrases in $r^Y$ with their synonyms in plain language. That is, for each $s^Y$ in $r^Y$, we substitute $s^Y$ with the synonymous $s^X$. The resulting synonymous response $r'$ is referred to as the modified response. Such direct substitution ensures the synonymity of $r'$ and $r^Y$. Note that $r'$ could be ungrammatical. This is acceptable because $r'$ is only used internally by the dialogue system to match with user input and is not shown to users.

\subsection{Context Rewriting}
\label{sec:main-corpus-context}

Now we have pairs of synonymous response $r'$ and stylized response $r^Y$, and a remaining problem is that the $r'$ and $r^Y$ may be totally irrelevant to the original context $c$ after style transfer.
So we attempt to modify $c$ to a more relevant one, $c'$, regarded as the stylized context.
To obtain $c'$, we select keywords similar to the stylized phrase $s^Y$ to append to $c$. For example, given $s^y$=``being multi-threaded'', the stylized context $c'$ could be [$c$ + ``simultaneously'' + ``moreover'']. We considered several choices of the keywords, including frequently co-occurred words, nearest words in word embedding space, and simply $s^X$ itself. We found fastText embedding~\shortcite{bojanowski2017enriching} yields more reasonable keywords. We choose the top 5 words with the highest cosine similarity to the stylized phrase $s^Y$.

\section{Evaluation}
\label{sec:eval}
We conducted both offline experiment and online A/B test to evaluate the performance of our framework. Each system is given an utterance in the testing set as input. The output response is evaluated in terms of its quality. 

\subsection{Dataset\footnote{All datasets will be released.}}
\label{Sec:eval-dataset}

To train a stylized dialogue system, our framework does not need paralleled data. Instead, we need a generic dialogue corpus $\mathcal{X}$, a stylized non-conversational corpus $\mathcal{T}$, and stylized jargon phrase - synonymous pairs.

\paragraph{Source dialogue corpus:} We use the dialogue corpus collected by \shortcite{shang2015neural} from Sina Weibo, a popular twitter-like service in China. The contexts are extracted from posts. Comments on the posts are taken as responses. The corpus covers various topics from different users, so we consider it to be in a generic language style. We normalized the tokens and removed non-content words like ``retweeted from''. Contexts and responses were tokenized by Jieba\footnote{https://github.com/fxsjy/jieba} toolkit. We sampled 100 posts for testing, and took the remaining 4.4 million context-response pairs as the source dialogue corpus.

\paragraph{Stylized non-conversational corpus:} Stylized non-conversational corpus are used for Response Filtering only. For each stylized phrase, we searched Sina Weibo and retrieved the first 500 matched posts. Comments were not retrieved. The posts are pre-processed similarly to the source dialogue corpus. We collected five corpora for the Internet slang style and the computer style.

\paragraph{Jargon phrase - synonymous pairs:} As mentioned before, We collected on average 523 parallel jargon phrase - synonym pairs per style.

\begin{table*}
\centering
\small
\begingroup
\setlength{\tabcolsep}{2.9pt} 
\begin{tabular}{l|R{0.12in}R{0.12in}R{0.12in}R{0.16in}|R{0.12in}R{0.12in}R{0.12in}R{0.16in}|R{0.12in}R{0.12in}R{0.12in}R{0.16in}|R{0.12in}R{0.12in}R{0.12in}R{0.16in}|R{0.12in}R{0.12in}R{0.12in}R{0.16in}||R{0.18in}R{0.19in}R{0.14in}R{0.14in}|R{0.2in}}
& \multicolumn{4}{c|}{Trendy Teenager} & \multicolumn{4}{c|}{Computer} & \multicolumn{4}{c|}{Physics} & \multicolumn{4}{c|}{Pok\'emon} & \multicolumn{4}{c||}{Martial Arts} & \multicolumn{5}{c}{Average} \\
& R & S & D1 & D2 & R & S & D1 & D2 & R & S & D1 & D2 & R & S & D1 & D2 & R & S & D1 & D2 & R & S & D1 & D2 & RSA \\
\hline
Generic & 1.3 & 0.7 & .51 & .84 & 1.6 & 0.1 & .51 & .84 & 1.6 & 0.0 & .51 & .84 & 1.4 & 0.0 & .51 & .84 & 1.3 & 0.1 & .51 & .84 & 1.45 & 0.17 & .51 & .84 & 0.81 \\
\hline
MTask+MMI & \textbf{1.1} & 0.7 & .19 & .40 & 1.0 & 0.1 & .19 & .38 & \textbf{1.1} & 0.0 & .21 & .40 & \textbf{1.0} & 0.0 & .19 & .39 & 1.1 & 0.0 & .19 & .37 & \textbf{1.07} & 0.16 & .19 & .39 & 0.62\\
CAAE-direct & 0.6 & 1.6 & .28 & .48 & 0.1 & 1.7 & .19 & .41 & 0.2 & 1.0 & .25 & .53 & 0.4 & 1.5 & .27 & .51 & 0.3 & 1.1 & .25 & .48 & 0.32 & 1.36 & .25 & .48 & 0.84 \\
CAAE-patch & 0.8 & 1.0 & .32 & .51 & 0.2 & 0.5 & .15 & .30 & 0.3 & 0.1 & .21 & .43 & 0.4 & 0.3 & .27 & .50 & 0.4 & 0.3 & .21 & .50 & 0.42 & 0.45 & .23 & .45 & 0.44 \\
DelRetGen-direct & 0.9 & \textbf{1.7} & .41 & .77 & 0.3 & \textbf{1.9} & .38 & .74 & 0.5 & 1.2 & .47 & .80 & 0.7 & 1.6 & .45 & .77 & 0.7 & \textbf{1.3} & .43 & .76 & 0.62 & \textbf{1.53} & .43 & .77 & 1.08 \\
Ours-direct & \textbf{1.1} & 1.5 & \textbf{.55} & \textbf{.84} & \textbf{1.1} & 0.9 & \textbf{.55} & \textbf{.84} & 1.0 & 0.8 & \textbf{.56} & \textbf{.84} & \textbf{1.0} & 1.2 & \textbf{.57} & \textbf{.83} & \textbf{0.9} & 1.0 & \textbf{.56} & \textbf{.84} & 1.00 & 1.10 & \textbf{.56} & \textbf{.84} & 1.05 \\
Ours-patch & \textbf{1.1} & \textbf{1.7} & .50 & .81 & 0.7 & 1.8 & .47 & .81 & 0.9 & \textbf{1.3} & .49 & .83 & 0.9 & \textbf{1.7} & .50 & \textbf{.83} & 0.8 & 1.2 & .50 & \textbf{.84} & 0.86 & \textbf{1.53} & .49 & .82 & \textbf{1.20} \\
\end{tabular}
\endgroup
\caption{Evaluation results of stylized dialogue. R: relevance, S: style, D1: distinct-1, D2: distinct-2, *-m: simplified variant without modifying response or context}
\label{tab:eval-dialogue}
\vspace{-9pt}
\end{table*}

\subsection{Experiment Settings}

\paragraph{Generic dialogue system:} We built a retrieval-based dialogue system as the start point for mimicking different styles. We use Elasticsearch~\shortcite{gormley2015elasticsearch} to build a conversation repository containing the source dialogue corpus $\mathcal{X}$. Given an input utterance, we retrieve the top 100 matched contexts and their paired responses. A pre-trained dialogue model selects a proper response as output. The dialogue model was trained with a different corpus other than the corpus $\mathcal{X}$ used in this paper.

\paragraph{Our approach:}
We ``patch'' the generic dialogue system with the proposed method. For each $(c, r)$ pair in $\mathcal{X}$, we try to rewrite it to obtain stylized responses. If the rewriting fails, we simply copy the pair, i.e., let $c'=c$ and $r^Y=r'=r$.

For ablation study, we consider a simplified variant, where the generic dialogue system is not patched and its output response $r$ is directly rewritten as $r^Y$. 
We denote our approach and the simplified variant as \textbf{Ours-patch} and \textbf{Ours-direct}.

\paragraph{Baselines:}
We compare with a generation-based approach by \citeyear{luan2017multi}. It uses multi-task learning to personalize a dialogue model. We train the model with our dialogue corpus $\mathcal{X}$ and non-conversational corpus $\mathcal{T}$. We improve its diversity by interpolating the MMI model by \citeyear{li2016diversity}. We set its encoder and decoder to be 2-layer LSTMs with 500 hidden units. Its vocabulary (60k) covers words in $\mathcal{X}$ and $\mathcal{T}$. We refer to this approach as \textbf{MTask+MMI}.

We also consider leveraging text style transfer models. Given an output response $r$ of the generic dialogue system, we use a style transfer model to transform $r$ to a stylized response. Two representative models are used, the cross-aligned auto-encoder by \citeyear{shen2017style} and the ``delete, retrieve, and generate'' model by \citeyear{li2018delete}. We denote the two approaches as \textbf{CAAE-direct} and \textbf{DelRetGen-direct}. In addition, we consider integrating the style transfer models into our framework. We replace the procedures of \textbf{Response Rewriting} with one of the style transfer models. We keep the remaining procedures and ``patch'' the generic dialogue system accordingly. This approach is denoted as \textbf{CAAE-patch}\footnote{We tried to optimize the original implementation of DelRetGen, but still can not rewrite the corpus in a reasonable time.}.

\subsection{Metrics}

We consider three criteria for response's quality: relevance, style degree, and content diversity.

\paragraph{Relevance:} Relevance refers to the extent of a response being proper for an input utterance. We follow the same guideline as \citeyear{shang2015neural} to rate Relevance on a scale of 0 to 2.


\paragraph{Style degree:} style degree refers to how likely the specified language style can be identified. We rate style degree on a scale of 0 to 2.

\begin{itemize}
    \item \textbf{Strong style (2):} Obvious usage of words in the specified language style can be found.
    \item \textbf{Weak style (1):} Two cases are included: (a) Stylized words are used, but it is still similar to everyday conversation, e.g., ``What's the name of this \textbf{software}?'' (b) Common words are used, but they have special meaning for the target persona. For example, ``I have an \textbf{abstract} girlfriend'', where ``abstract'' has a specific meaning in programming.
    \item \textbf{No style (0):} The response is not fluent or does dot show any identifiable style.
\end{itemize}

\paragraph{Diversity:} Diversity of response is recognized as an important metric of dialogue systems~\shortcite{li2016diversity}. Dialogue systems with good diversity produce more interesting responses on various topics. We measure the diversity of responses by the number of distinct unigrams and bigrams divided by total number of words in the responses, i.e., Distinct-1 (D1) and Distinct-2 (D2).

For each style, we obtained 700 responses from the 100 testing input utterances by the evaluated systems. An input utterance and a shuffled list of responses were presented to 3 annotators each time. The annotators were asked to rate each response according to the above guideline.

To measure agreement, we use a generalized form of Scott's pi coefficient with linear weighting~\shortcite{gwet2014handbook}. The coefficients indicate fair agreement among ratings (0.37 and 0.44 for relevance; 0.56 and 0.79 for style degree).

\begin{figure}
\centering
\small
\begingroup
\setlength{\tabcolsep}{3pt}
\begin{tabular}{l|p{2.2in}}
\multirow{2}{*}{User Query} & \chinese{好久没吃了，好想吃哇} \\
& Haven't eat it for a long time and really want to eat it. \\
\hline
\multirow{2}{*}{Trendy Teenagers} & \chinese{好久没吃了啊，减肥\underline{不存在的}，能不诱惑我吗？} \\
& Haven't eat it for a lone time. Losing weight is \textbf{non-existent}. Could you not tempt me? \\
\hline
\multirow{2}{*}{Computer Junkies} & \chinese{想吃，\underline{贪婪算法}告诉我怎么吃吗？}\\
& Want to eat. Would \textbf{greedy algorithms} tell me how to eat? \\
\hline
\multirow{2}{*}{Physics Fans} & \chinese{好饿啊。看到这汉堡就控制\underline{运动速度}。} \\
& So hungry. I have to control the \textbf{movement velocity} when I see this hamburger.\\
\hline
\multirow{2}{*}{Pok\'emon Fans} & \chinese{哈哈，我爱\underline{舔舌}头吃！}\\
& Haha, I love eating by \textbf{lick}. \\
\hline
\multirow{2}{*}{Martial Arts Fans} & \chinese{不是好久没吃过，\underline{苏荃}从来都没吃过…} \\
& Not for a long time, \textbf{Su Quan}\footnotemark has never eaten it. \\

\hline
\end{tabular}
\endgroup
\caption{Sample responses by our approach.}
\label{fig:sample-dialogue}
\end{figure}

\subsection{Evaluation Result}
Table \ref{tab:eval-dialogue} shows evaluation results of stylized dialogue. The average of relevance and style degree ratings (denoted as RSA) is included for overall performance. The multi-task approach \textbf{MTask+MMI} does not produce stylized response as expected. We attribute it to the low overlap between the dialogue corpus and the target corpus. On average, only 81.4\% words of the target corpus (excluding stop words) appear in the vocabulary of dialogue corpus. The decoder fails to correlate non-overlapping words with the utterance's encoding, even after we extended its vocabulary.

For CAAE-patch, its relevance increases after modifying context and response, but its style degree decreases dramatically. The average lengths of response are respectively 7.4 and 5.7 tokens for CAAE-direct and CAAE-patch. The root cause for this behavior is the low content diversity. The model tends to produce short and common responses. The dialogue model has to choose a ``safe'' but more relevant one, when no other relevant and informative response is available. 

Our simplified approach (Ours-direct) achieves a higher relevance and a lower style degree than our complete approach (ours-patch). Among the responses produced by the base system, the simplified approach fails to transfer any response for 25\% of queries, whereas ours-patch succeed to return at least one stylized responses for all queries. If we count the queries on which both approaches succeed,
the relevance is 0.78 and 0.86, and the style degree is 1.43 and 1.52, for the simplified and the complete approaches. The diversity remains similar. This strongly supports our idea of distinguishing transferring style from keeping meaning unchanged does work well. In general, our complete approach (ours-patch) achieves the best overall performance in terms of RSA (t-tests with $p<10^{-4})$. Figure \ref{fig:sample-dialogue} illustrates our approach's responses.

\footnotetext{A character in the novel series \textit{The Legend of Lu Xiaofeng}.}

\begin{figure*}[h]
\centering
\includegraphics[width=6.25in]{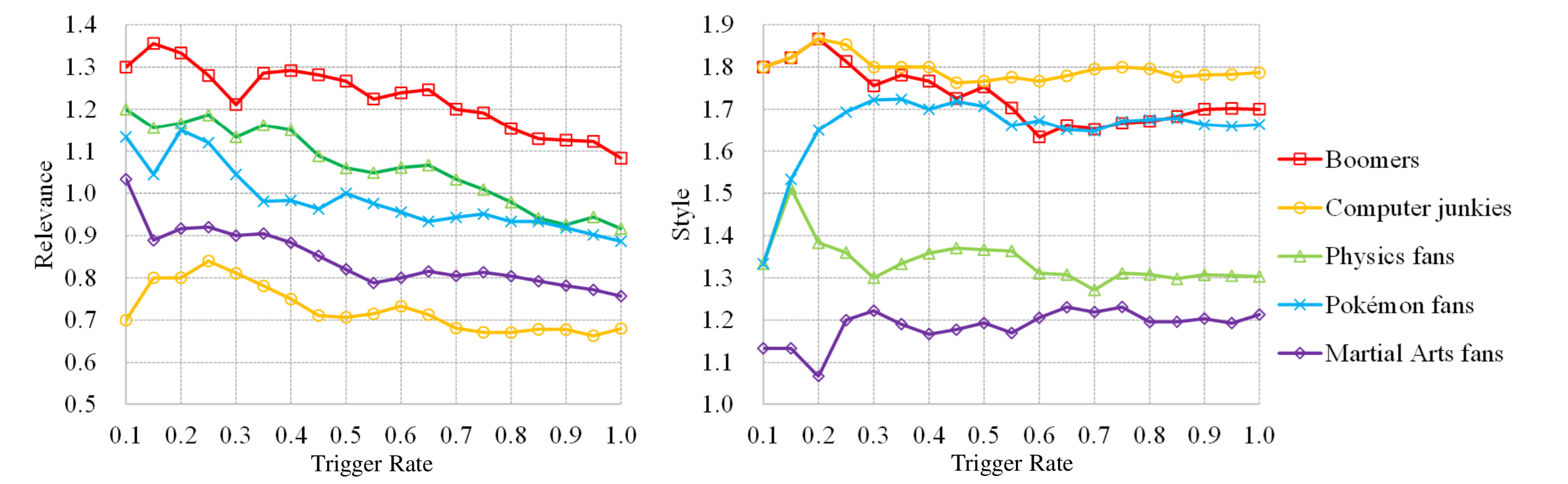}
\caption{Relevance and style degree change when we trigger different ratios of stylized responses}
\label{fig:corpus-coverage}
\end{figure*}

\subsection{Changes with Trigger Rate}
There is a trade-off between style degree and relevance. When we allow more stylized responses to return during conversations, a chatbot has more chance to exhibit as its style. However, there is a risk of lowering the relevance. Thus we use trigger rate as our variant and draw curves to show how relevance and style degree change at the same time.

Based on our human labeled data, we simulate the results when different trigger rates are applied. For example, given a style, e.g., trendy teenagers, we set a threshold of the ensemble regression model output to filter out 90\% stylized responses. Thus we get 10\% trigger rate. Then we average the relevance score and the style degree as the corresponding evaluation results, e.g., 1.3 and 1.8 for the trendy teenager style. Similarly, we can filter out different ratio of stylized responses and get relevance and style degree scores at different trigger rate. Figure \ref{fig:corpus-coverage} shows us the curves for five styles. 

Take the trendy teenager style as an example. We observe that relevance remains high until about 15\% stylized responses are triggered. Then relevance goes down gradually. Even if all stylized responses are triggered, the relevance score is around 1.1, which means above fair. The style curve has a similar trend and the highest style degree is achieved when 20\% stylized responses are triggered. At that position, the relevance score is close to the maximum. Therefore, if our goal is to add some sense of style while keeping high relevance, we can choose the trigger rate of 20\%. It means that in almost one fifth turns users can get the stylized responses. That is a proper ratio as even a real trendy teenager does not show her/his style at all utterances.

As Figure \ref{fig:corpus-coverage} shows, other styles have similar trends. We can find a proper trade-off between relevance and style degree in the curves.
Our approach provides the flexibility to adjust a dialogue system according to the requirements of relevance and style degree. Some styles may be more difficult than some other styles to transfer. For example, the computer style has lowest relevance. Although the style degree is as high as 1.8, the relevance score is only 0.7 when the trigger rate is 10\%. This indicate that there is a big gap between the dialogue corpus and the stylized non-conversational corpus for the computer style. We guess that Weibo users like publishing content about their life, more than about techniques. 

\begin{table}
\centering
\begingroup
\begin{tabular}{l|rrrr|r}
& \multicolumn{1}{c}{F} & \multicolumn{1}{c}{S} & \multicolumn{1}{c}{D1} & \multicolumn{1}{c|}{D2} & \multicolumn{1}{c}{FSA} \\
\hline
Source & 1.86 & 0.18 & .50 & .84 & 1.02 \\
\hline
CAAE & 1.09 & 1.30 & .14 & .32 & 1.20 \\
DelRetGen & 1.01 & \textbf{1.41} & .41 & .77 & 1.21 \\
 Ours & \textbf{1.43} & 1.19 & \textbf{.53} & \textbf{.84} & \textbf{1.31} \\
\end{tabular}
\endgroup
\caption{Evaluation results of transferred responses by single models. LM: Language Model}
\label{tab:eval-corpus-single}
\vspace{-9pt}
\end{table}

\subsection{Analysis of Style Transfer}
In this study, we propose a first study to transfer a generic corpus to a stylized one. To evaluate how effective our transfer approach is, we further compare our style transfer model with two baseline methods, \textbf{CAAE} and \textbf{DelRetGen}. The evaluation metrics of style transfer are similar to that of stylized dialogue, and we only change the relevance metric to fluency metric. The fluency metric is rated on a scale of 0 to 2 that represent  completely non-sense (0), understandable with minor errors (1), and fluent without error (2), respectively.

Table \ref{tab:eval-corpus-single} shows the evaluation results. As expected, source responses are rated with perfect fluency and low style degree. They slightly exhibit Internet slang's style, because the dataset is collected from social media. On average, our approach achieves much better fluency and diversity but lower style degree compared to the two baselines. CAAE tends to generate ``safe'' sentences which make use of only a few target style phrases. To see this, we counted distinct target style phrases in transferred responses. The averaged counts are respectively 81, 172, and 181, for CAAE, DelRetGen, and our approach. DelRetGen improves diversity by involving more target style phrases than CAAE. However, its generation step tends to generate frequent words, limiting its overall diversity. In general, our approach outperforms the baselines on overall performance (FSA, D1, and D2; t-tests with $p < 10^{-4}$). 

\subsection{Online Evaluation}
Compared to the generic dialogue system, our framework (as well as other stylized systems) produces more stylized but less relevant responses. We wonder if it has a positive or negative impact on user satisfaction. We conducted A/B test on a commercial social chatbot during one month. We select the trendy teenager style as our testbed.

We randomly selected a fraction of users as the treatment group. They were served with the patched system. In case where the patched system fails to find a relevant response, we fallback to the generic system. Users in the control group were always served with the generic system. After the A/B test, we collected 12K conversations for the treatment group. A conversation is a sequence $\{(u_1, r_1), (u_2, r_2)\}$, where $u$ is a user utterance and $r$ is a response. We let $r_1$ be the patched system's response. The same number of conversations were sampled from the control group.

We consider two automatic metrics to measure user satisfaction. The first metric is the sentiment of user's follow-up utterance. Positive sentiment reflects better user satisfaction. We use a sentiment classifier~\cite{zeng2019attitude} to classify the follow-up utterance $u_2$. The classifier outputs labels of $\{-1,0,1\}$, for negative, neutral, and positive sentiment. We take the average label for each group as \textit{Sentiment} in Table \ref{tab:online-eval}. The second metric is the fraction of overlapping words between $r_1$ and $u_2$ (See \textit{Automatic follow-up} in Table \ref{tab:online-eval}). It indicates whether the user is following $r_1$ or starting a new topic.

We also use human annotation for evaluation. We sampled 250 conversations from each group, and asked 5 human annotators to rate user's satisfaction on a scale of -2 to +2:
\begin{itemize}
    \item \textbf{Satisfied (+2)} The user is satisfied with the conversation and shows positive emotion. 
    \item \textbf{Follow (+1)} The user understands the response $r_1$ and follows the topic.
    \item \textbf{Neutral (0)} The user keeps chatting but to ignore $r_1$.
    \item \textbf{Not-follow (-1)} The user is confused about $r_1$ or simply exits the chatbot.
    \item \textbf{Unsatisfied (-2)} The user is upset about the conversation and shows negative emotion.
\end{itemize}

Table \ref{tab:online-eval} shows the comparison in the above metrics. The patched system performs consistently better than the baseline system across all metrics.

\begin{table}
\centering
\begin{tabular}{c|cc}
& Treatment & Control   \\
\hline
Sentiment & \textbf{0.0250} & 0.0217  \\
Automatic follow-up & \textbf{0.1120} & 0.10445 \\
Human rating & \textbf{0.488} & 0.436  \\
\end{tabular}
\caption{Evaluation result of online A/B test.}
\label{tab:online-eval}
\vspace{-6pt}
\end{table}

\section{Conclusion and Future Work}
\label{sec:conclude}
We propose a light-weight ``patching'' framework that enables a retrieval-based dialogue system to mimic the language style of a specified character setting. Our framework is flexible and less expensive to apply. Our approach is based on transforming a generic dialogue corpus to a stylized corpus. We design a series of strategies to implement the transformation. 
Five character settings are implemented. The experiments show that, with several hundreds of manually annotated jargon phrases, our approach is able to produce fluent, relevant, stylized, and diverse responses. A/B test indicates that users are more satisfied about the stylized system.

As  future work, we plan to improve relevance further and experiment with combining two or more styles in one chatbot. Theoretically our proposed framework is additive, because it provides a way to put different but non-contradictory personalities together. For example, a chatbot could be a computer junkie and a martial arts fan at the same time. When both character settings have response candidates, we can choose one of them by pre-defined priors. It would be interesting to implement such combination and test its effectiveness. 

\bibliographystyle{ACM-Reference-Format}
\bibliography{reference}


\begin{thebibliography}{44}


\ifx \showCODEN    \undefined \def \showCODEN     #1{\unskip}     \fi
\ifx \showDOI      \undefined \def \showDOI       #1{#1}\fi
\ifx \showISBNx    \undefined \def \showISBNx     #1{\unskip}     \fi
\ifx \showISBNxiii \undefined \def \showISBNxiii  #1{\unskip}     \fi
\ifx \showISSN     \undefined \def \showISSN      #1{\unskip}     \fi
\ifx \showLCCN     \undefined \def \showLCCN      #1{\unskip}     \fi
\ifx \shownote     \undefined \def \shownote      #1{#1}          \fi
\ifx \showarticletitle \undefined \def \showarticletitle #1{#1}   \fi
\ifx \showURL      \undefined \def \showURL       {\relax}        \fi
\providecommand\bibfield[2]{#2}
\providecommand\bibinfo[2]{#2}
\providecommand\natexlab[1]{#1}
\providecommand\showeprint[2][]{arXiv:#2}

\bibitem[\protect\citeauthoryear{Bojanowski, Grave, Joulin, and
  Mikolov}{Bojanowski et~al\mbox{.}}{2017}]%
        {bojanowski2017enriching}
\bibfield{author}{\bibinfo{person}{Piotr Bojanowski}, \bibinfo{person}{Edouard
  Grave}, \bibinfo{person}{Armand Joulin}, {and} \bibinfo{person}{Tomas
  Mikolov}.} \bibinfo{year}{2017}\natexlab{}.
\newblock \showarticletitle{Enriching word vectors with subword information}.
\newblock \bibinfo{journal}{\emph{Transactions of the Association for
  Computational Linguistics}}  \bibinfo{volume}{5} (\bibinfo{year}{2017}).
\newblock


\bibitem[\protect\citeauthoryear{Devlin, Chang, Lee, and Toutanova}{Devlin
  et~al\mbox{.}}{2019}]%
        {devlin2019BERT}
\bibfield{author}{\bibinfo{person}{Jacob Devlin}, \bibinfo{person}{Ming-Wei
  Chang}, \bibinfo{person}{Kenton Lee}, {and} \bibinfo{person}{Kristina
  Toutanova}.} \bibinfo{year}{2019}\natexlab{}.
\newblock \showarticletitle{BERT: Pre-training of Deep Bidirectional
  Transformers for Language Understanding}. In
  \bibinfo{booktitle}{\emph{Proceedings of the 2019 Conference of the North
  American Chapter of the Association for Computational Linguistics: Human
  Language Technologies}}.
\newblock


\bibitem[\protect\citeauthoryear{Fu, Zhou, Chen, and Li}{Fu
  et~al\mbox{.}}{2019}]%
        {fu2019rethinking}
\bibfield{author}{\bibinfo{person}{Yao Fu}, \bibinfo{person}{Hao Zhou},
  \bibinfo{person}{Jiaze Chen}, {and} \bibinfo{person}{Lei Li}.}
  \bibinfo{year}{2019}\natexlab{}.
\newblock \showarticletitle{Rethinking Text Attribute Transfer: A Lexical
  Analysis}.
\newblock  (\bibinfo{year}{2019}).
\newblock


\bibitem[\protect\citeauthoryear{Fu, Tan, Peng, Zhao, and Yan}{Fu
  et~al\mbox{.}}{2018}]%
        {fu2018style}
\bibfield{author}{\bibinfo{person}{Zhenxin Fu}, \bibinfo{person}{Xiaoye Tan},
  \bibinfo{person}{Nanyun Peng}, \bibinfo{person}{Dongyan Zhao}, {and}
  \bibinfo{person}{Rui Yan}.} \bibinfo{year}{2018}\natexlab{}.
\newblock \showarticletitle{Style transfer in text: Exploration and
  evaluation}. In \bibinfo{booktitle}{\emph{The 32nd AAAI Conference on
  Artificial Intelligence}}.
\newblock


\bibitem[\protect\citeauthoryear{Gormley and Tong}{Gormley and Tong}{2015}]%
        {gormley2015elasticsearch}
\bibfield{author}{\bibinfo{person}{Clinton Gormley} {and}
  \bibinfo{person}{Zachary Tong}.} \bibinfo{year}{2015}\natexlab{}.
\newblock \bibinfo{booktitle}{\emph{Elasticsearch: the definitive guide: a
  distributed real-time search and analytics engine}}.
\newblock \bibinfo{publisher}{" O'Reilly Media, Inc."}.
\newblock


\bibitem[\protect\citeauthoryear{Goyal, Srinivasan, Anandhavelu, and
  Sancheti}{Goyal et~al\mbox{.}}{2021}]%
        {goyal2021multi}
\bibfield{author}{\bibinfo{person}{Navita Goyal}, \bibinfo{person}{Balaji~Vasan
  Srinivasan}, \bibinfo{person}{N Anandhavelu}, {and}
  \bibinfo{person}{Abhilasha Sancheti}.} \bibinfo{year}{2021}\natexlab{}.
\newblock \showarticletitle{Multi-Style Transfer with Discriminative Feedback
  on Disjoint Corpus}. In \bibinfo{booktitle}{\emph{Proceedings of the 2021
  Conference of the North American Chapter of the Association for Computational
  Linguistics: Human Language Technologies}}. \bibinfo{pages}{3500--3510}.
\newblock


\bibitem[\protect\citeauthoryear{Gwet}{Gwet}{2014}]%
        {gwet2014handbook}
\bibfield{author}{\bibinfo{person}{Kilem~L Gwet}.}
  \bibinfo{year}{2014}\natexlab{}.
\newblock \bibinfo{booktitle}{\emph{Handbook of inter-rater reliability: The
  definitive guide to measuring the extent of agreement among raters}}.
\newblock \bibinfo{publisher}{Advanced Analytics, LLC}.
\newblock


\bibitem[\protect\citeauthoryear{Herzig, Shmueli-Scheuer, Sandbank, and
  Konopnicki}{Herzig et~al\mbox{.}}{2017}]%
        {herzig2017neural}
\bibfield{author}{\bibinfo{person}{Jonathan Herzig}, \bibinfo{person}{Michal
  Shmueli-Scheuer}, \bibinfo{person}{Tommy Sandbank}, {and}
  \bibinfo{person}{David Konopnicki}.} \bibinfo{year}{2017}\natexlab{}.
\newblock \showarticletitle{Neural response generation for customer service
  based on personality traits}. In \bibinfo{booktitle}{\emph{Proceedings of the
  10th International Conference on Natural Language Generation}}.
\newblock


\bibitem[\protect\citeauthoryear{Hu, Yang, Liang, Salakhutdinov, and Xing}{Hu
  et~al\mbox{.}}{2017}]%
        {hu2017toward}
\bibfield{author}{\bibinfo{person}{Zhiting Hu}, \bibinfo{person}{Zichao Yang},
  \bibinfo{person}{Xiaodan Liang}, \bibinfo{person}{Ruslan Salakhutdinov},
  {and} \bibinfo{person}{Eric~P Xing}.} \bibinfo{year}{2017}\natexlab{}.
\newblock \showarticletitle{Toward controlled generation of text}. In
  \bibinfo{booktitle}{\emph{Proceedings of the 34th International Conference on
  Machine Learning}}.
\newblock


\bibitem[\protect\citeauthoryear{Jena, Vashisht, Basu, Ungar, and Sedoc}{Jena
  et~al\mbox{.}}{2017}]%
        {jena2017enterprise}
\bibfield{author}{\bibinfo{person}{Grishma Jena}, \bibinfo{person}{Mansi
  Vashisht}, \bibinfo{person}{Abheek Basu}, \bibinfo{person}{Lyle Ungar}, {and}
  \bibinfo{person}{Joao Sedoc}.} \bibinfo{year}{2017}\natexlab{}.
\newblock \showarticletitle{Enterprise to computer: Star trek chatbot}.
\newblock \bibinfo{journal}{\emph{arXiv preprint arXiv:1708.00818}}
  (\bibinfo{year}{2017}).
\newblock


\bibitem[\protect\citeauthoryear{John, Mou, Bahuleyan, and Vechtomova}{John
  et~al\mbox{.}}{2019}]%
        {john2019disentangled}
\bibfield{author}{\bibinfo{person}{Vineet John}, \bibinfo{person}{Lili Mou},
  \bibinfo{person}{Hareesh Bahuleyan}, {and} \bibinfo{person}{Olga
  Vechtomova}.} \bibinfo{year}{2019}\natexlab{}.
\newblock \showarticletitle{Disentangled Representation Learning for
  Non-Parallel Text Style Transfer}. In \bibinfo{booktitle}{\emph{Proceedings
  of the 57th Annual Meeting of the Association for Computational
  Linguistics}}.
\newblock


\bibitem[\protect\citeauthoryear{Krishna, Wieting, and Iyyer}{Krishna
  et~al\mbox{.}}{2020}]%
        {krishna2020reformulating}
\bibfield{author}{\bibinfo{person}{Kalpesh Krishna}, \bibinfo{person}{John
  Wieting}, {and} \bibinfo{person}{Mohit Iyyer}.}
  \bibinfo{year}{2020}\natexlab{}.
\newblock \showarticletitle{Reformulating Unsupervised Style Transfer as
  Paraphrase Generation}. In \bibinfo{booktitle}{\emph{Proceedings of the 2020
  Conference on Empirical Methods in Natural Language Processing}}.
\newblock


\bibitem[\protect\citeauthoryear{Li, Galley, Brockett, Gao, and Dolan}{Li
  et~al\mbox{.}}{2016a}]%
        {li2016diversity}
\bibfield{author}{\bibinfo{person}{Jiwei Li}, \bibinfo{person}{Michel Galley},
  \bibinfo{person}{Chris Brockett}, \bibinfo{person}{Jianfeng Gao}, {and}
  \bibinfo{person}{Bill Dolan}.} \bibinfo{year}{2016}\natexlab{a}.
\newblock \showarticletitle{A Diversity-Promoting Objective Function for Neural
  Conversation Models}. In \bibinfo{booktitle}{\emph{Proceedings of the 2016
  Conference of the North American Chapter of the Association for Computational
  Linguistics: Human Language Technologies}}.
\newblock


\bibitem[\protect\citeauthoryear{Li, Galley, Brockett, Spithourakis, Gao, and
  Dolan}{Li et~al\mbox{.}}{2016b}]%
        {li2016persona}
\bibfield{author}{\bibinfo{person}{Jiwei Li}, \bibinfo{person}{Michel Galley},
  \bibinfo{person}{Chris Brockett}, \bibinfo{person}{Georgios Spithourakis},
  \bibinfo{person}{Jianfeng Gao}, {and} \bibinfo{person}{Bill Dolan}.}
  \bibinfo{year}{2016}\natexlab{b}.
\newblock \showarticletitle{A Persona-Based Neural Conversation Model}. In
  \bibinfo{booktitle}{\emph{Proceedings of the 54th Annual Meeting of the
  Association for Computational Linguistics}}.
\newblock


\bibitem[\protect\citeauthoryear{Li, Jia, He, and Liang}{Li
  et~al\mbox{.}}{2018}]%
        {li2018delete}
\bibfield{author}{\bibinfo{person}{Juncen Li}, \bibinfo{person}{Robin Jia},
  \bibinfo{person}{He He}, {and} \bibinfo{person}{Percy Liang}.}
  \bibinfo{year}{2018}\natexlab{}.
\newblock \showarticletitle{Delete, Retrieve, Generate: a Simple Approach to
  Sentiment and Style Transfer}. In \bibinfo{booktitle}{\emph{Proceedings of
  the 2018 Conference of the North American Chapter of the Association for
  Computational Linguistics: Human Language Technologies}}.
\newblock


\bibitem[\protect\citeauthoryear{Li, Li, Mou, Jiang, Lyu, and King}{Li
  et~al\mbox{.}}{2020}]%
        {li2020unsupervised}
\bibfield{author}{\bibinfo{person}{Jingjing Li}, \bibinfo{person}{Zichao Li},
  \bibinfo{person}{Lili Mou}, \bibinfo{person}{Xin Jiang},
  \bibinfo{person}{Michael~R Lyu}, {and} \bibinfo{person}{Irwin King}.}
  \bibinfo{year}{2020}\natexlab{}.
\newblock \showarticletitle{Unsupervised Text Generation by Learning from
  Search}. In \bibinfo{booktitle}{\emph{Advances in neural information
  processing systems}}.
\newblock


\bibitem[\protect\citeauthoryear{Liu, Neubig, and Wieting}{Liu
  et~al\mbox{.}}{2021}]%
        {liu2021learning}
\bibfield{author}{\bibinfo{person}{Yixin Liu}, \bibinfo{person}{Graham Neubig},
  {and} \bibinfo{person}{John Wieting}.} \bibinfo{year}{2021}\natexlab{}.
\newblock \showarticletitle{On Learning Text Style Transfer with Direct
  Rewards}. In \bibinfo{booktitle}{\emph{Proceedings of the 2021 Conference of
  the North American Chapter of the Association for Computational Linguistics:
  Human Language Technologies}}. \bibinfo{pages}{4262--4273}.
\newblock


\bibitem[\protect\citeauthoryear{Luan, Brockett, Dolan, Gao, and Galley}{Luan
  et~al\mbox{.}}{2017}]%
        {luan2017multi}
\bibfield{author}{\bibinfo{person}{Yi Luan}, \bibinfo{person}{Chris Brockett},
  \bibinfo{person}{Bill Dolan}, \bibinfo{person}{Jianfeng Gao}, {and}
  \bibinfo{person}{Michel Galley}.} \bibinfo{year}{2017}\natexlab{}.
\newblock \showarticletitle{Multi-Task Learning for Speaker-Role Adaptation in
  Neural Conversation Models}. In \bibinfo{booktitle}{\emph{Proceedings of the
  8th International Joint Conference on Natural Language Processing}}.
\newblock


\bibitem[\protect\citeauthoryear{Luo, Huang, Zeng, Nie, and Sun}{Luo
  et~al\mbox{.}}{2019}]%
        {luo2019learning}
\bibfield{author}{\bibinfo{person}{Liangchen Luo}, \bibinfo{person}{Wenhao
  Huang}, \bibinfo{person}{Qi Zeng}, \bibinfo{person}{Zaiqing Nie}, {and}
  \bibinfo{person}{Xu Sun}.} \bibinfo{year}{2019}\natexlab{}.
\newblock \showarticletitle{Learning personalized end-to-end goal-oriented
  dialog}. In \bibinfo{booktitle}{\emph{Proceedings of the AAAI Conference on
  Artificial Intelligence}}.
\newblock


\bibitem[\protect\citeauthoryear{Madotto, Lin, Wu, and Fung}{Madotto
  et~al\mbox{.}}{2019}]%
        {madotto2019personalizing}
\bibfield{author}{\bibinfo{person}{Andrea Madotto}, \bibinfo{person}{Zhaojiang
  Lin}, \bibinfo{person}{Chien-Sheng Wu}, {and} \bibinfo{person}{Pascale
  Fung}.} \bibinfo{year}{2019}\natexlab{}.
\newblock \showarticletitle{Personalizing dialogue agents via meta-learning}.
  In \bibinfo{booktitle}{\emph{Proceedings of the 57th Annual Meeting of the
  Association for Computational Linguistics}}.
\newblock


\bibitem[\protect\citeauthoryear{Niu and Bansal}{Niu and Bansal}{2018}]%
        {niu2018polite}
\bibfield{author}{\bibinfo{person}{Tong Niu} {and} \bibinfo{person}{Mohit
  Bansal}.} \bibinfo{year}{2018}\natexlab{}.
\newblock \showarticletitle{Polite Dialogue Generation Without Parallel Data}.
\newblock \bibinfo{journal}{\emph{Transactions of the Association for
  Computational Linguistics}}  \bibinfo{volume}{6} (\bibinfo{year}{2018}).
\newblock


\bibitem[\protect\citeauthoryear{Shang, Lu, and Li}{Shang
  et~al\mbox{.}}{2015}]%
        {shang2015neural}
\bibfield{author}{\bibinfo{person}{Lifeng Shang}, \bibinfo{person}{Zhengdong
  Lu}, {and} \bibinfo{person}{Hang Li}.} \bibinfo{year}{2015}\natexlab{}.
\newblock \showarticletitle{Neural Responding Machine for Short-Text
  Conversation}. In \bibinfo{booktitle}{\emph{Proceedings of the 53rd Annual
  Meeting of the Association for Computational Linguistics and the 7th
  International Joint Conference on Natural Language Processing}}.
\newblock


\bibitem[\protect\citeauthoryear{Shen, Lei, Barzilay, and Jaakkola}{Shen
  et~al\mbox{.}}{2017}]%
        {shen2017style}
\bibfield{author}{\bibinfo{person}{Tianxiao Shen}, \bibinfo{person}{Tao Lei},
  \bibinfo{person}{Regina Barzilay}, {and} \bibinfo{person}{Tommi Jaakkola}.}
  \bibinfo{year}{2017}\natexlab{}.
\newblock \showarticletitle{Style transfer from non-parallel text by
  cross-alignment}. In \bibinfo{booktitle}{\emph{Advances in neural information
  processing systems}}.
\newblock


\bibitem[\protect\citeauthoryear{Shum, He, and Li}{Shum et~al\mbox{.}}{2018}]%
        {shum2018eliza}
\bibfield{author}{\bibinfo{person}{Heung-Yeung Shum},
  \bibinfo{person}{Xiao-dong He}, {and} \bibinfo{person}{Di Li}.}
  \bibinfo{year}{2018}\natexlab{}.
\newblock \showarticletitle{From Eliza to XiaoIce: challenges and opportunities
  with social chatbots}.
\newblock \bibinfo{journal}{\emph{Frontiers of Information Technology \&
  Electronic Engineering}} \bibinfo{volume}{19}, \bibinfo{number}{1}
  (\bibinfo{year}{2018}), \bibinfo{pages}{10--26}.
\newblock


\bibitem[\protect\citeauthoryear{Song, Wang, Zhang, Zhang, and Liu}{Song
  et~al\mbox{.}}{2021}]%
        {song2021bob}
\bibfield{author}{\bibinfo{person}{Haoyu Song}, \bibinfo{person}{Yan Wang},
  \bibinfo{person}{Kaiyan Zhang}, \bibinfo{person}{Wei-Nan Zhang}, {and}
  \bibinfo{person}{Ting Liu}.} \bibinfo{year}{2021}\natexlab{}.
\newblock \showarticletitle{BoB: BERT Over BERT for Training Persona-based
  Dialogue Models from Limited Personalized Data}.
\newblock \bibinfo{journal}{\emph{arXiv preprint arXiv:2106.06169}}
  (\bibinfo{year}{2021}).
\newblock


\bibitem[\protect\citeauthoryear{Song, Wang, Zhang, Liu, and Liu}{Song
  et~al\mbox{.}}{2020}]%
        {song2020generate}
\bibfield{author}{\bibinfo{person}{Haoyu Song}, \bibinfo{person}{Yan Wang},
  \bibinfo{person}{Wei-Nan Zhang}, \bibinfo{person}{Xiaojiang Liu}, {and}
  \bibinfo{person}{Ting Liu}.} \bibinfo{year}{2020}\natexlab{}.
\newblock \showarticletitle{Generate, delete and rewrite: A three-stage
  framework for improving persona consistency of dialogue generation}.
\newblock \bibinfo{journal}{\emph{arXiv preprint arXiv:2004.07672}}
  (\bibinfo{year}{2020}).
\newblock


\bibitem[\protect\citeauthoryear{Su, Cai, Wang, Baker, Korhonen, Collier, and
  Liu}{Su et~al\mbox{.}}{2020}]%
        {IGRL}
\bibfield{author}{\bibinfo{person}{Yixuan Su}, \bibinfo{person}{Deng Cai},
  \bibinfo{person}{Yan Wang}, \bibinfo{person}{Simon Baker},
  \bibinfo{person}{Anna Korhonen}, \bibinfo{person}{Nigel Collier}, {and}
  \bibinfo{person}{Xiaojiang Liu}.} \bibinfo{year}{2020}\natexlab{}.
\newblock \showarticletitle{Stylistic Dialogue Generation via
  Information-Guided Reinforcement Learning Strategy}.
\newblock \bibinfo{journal}{\emph{CoRR}}  \bibinfo{volume}{abs/2004.02202}
  (\bibinfo{year}{2020}).
\newblock
\showeprint[arxiv]{2004.02202}


\bibitem[\protect\citeauthoryear{Su, Yan, Cai, Baker, Korhonen, and Collier}{Su
  et~al\mbox{.}}{2021}]%
        {su2021prototype}
\bibfield{author}{\bibinfo{person}{Yixuan Su}, \bibinfo{person}{Wang Yan},
  \bibinfo{person}{Deng Cai}, \bibinfo{person}{Simon Baker},
  \bibinfo{person}{Anna Korhonen}, {and} \bibinfo{person}{Nigel Collier}.}
  \bibinfo{year}{2021}\natexlab{}.
\newblock \showarticletitle{Prototype-to-style: Dialogue generation with
  style-aware editing on retrieval memory}.
\newblock \bibinfo{journal}{\emph{IEEE/ACM Transactions on Audio, Speech, and
  Language Processing}} (\bibinfo{year}{2021}).
\newblock


\bibitem[\protect\citeauthoryear{Sudhakar, Upadhyay, and Maheswaran}{Sudhakar
  et~al\mbox{.}}{2019}]%
        {sudhakar2019transforming}
\bibfield{author}{\bibinfo{person}{Akhilesh Sudhakar}, \bibinfo{person}{Bhargav
  Upadhyay}, {and} \bibinfo{person}{Arjun Maheswaran}.}
  \bibinfo{year}{2019}\natexlab{}.
\newblock \showarticletitle{Transforming Delete, Retrieve, Generate Approach
  for Controlled Text Style Transfer}. In \bibinfo{booktitle}{\emph{Proceedings
  of the 2019 Conference on Empirical Methods in Natural Language Processing
  and the 9th International Joint Conference on Natural Language Processing}}.
\newblock


\bibitem[\protect\citeauthoryear{Tao, Wu, Xu, Hu, Zhao, and Yan}{Tao
  et~al\mbox{.}}{2019}]%
        {tao2019multi}
\bibfield{author}{\bibinfo{person}{Chongyang Tao}, \bibinfo{person}{Wei Wu},
  \bibinfo{person}{Can Xu}, \bibinfo{person}{Wenpeng Hu},
  \bibinfo{person}{Dongyan Zhao}, {and} \bibinfo{person}{Rui Yan}.}
  \bibinfo{year}{2019}\natexlab{}.
\newblock \showarticletitle{Multi-representation fusion network for multi-turn
  response selection in retrieval-based chatbots}. In
  \bibinfo{booktitle}{\emph{WSDM}}. \bibinfo{pages}{267--275}.
\newblock


\bibitem[\protect\citeauthoryear{Wu, Ren, Luo, and Sun}{Wu
  et~al\mbox{.}}{2019a}]%
        {wu2019hierarchical}
\bibfield{author}{\bibinfo{person}{Chen Wu}, \bibinfo{person}{Xuancheng Ren},
  \bibinfo{person}{Fuli Luo}, {and} \bibinfo{person}{Xu Sun}.}
  \bibinfo{year}{2019}\natexlab{a}.
\newblock \showarticletitle{A Hierarchical Reinforced Sequence Operation Method
  for Unsupervised Text Style Transfer}. In
  \bibinfo{booktitle}{\emph{Proceedings of the 57th Annual Meeting of the
  Association for Computational Linguistics}}.
\newblock


\bibitem[\protect\citeauthoryear{Wu, Zhang, Zang, Han, and Hu}{Wu
  et~al\mbox{.}}{2019b}]%
        {wu2019mask}
\bibfield{author}{\bibinfo{person}{Xing Wu}, \bibinfo{person}{Tao Zhang},
  \bibinfo{person}{Liangjun Zang}, \bibinfo{person}{Jizhong Han}, {and}
  \bibinfo{person}{Songlin Hu}.} \bibinfo{year}{2019}\natexlab{b}.
\newblock \showarticletitle{Mask and infill: applying masked language model to
  sentiment transfer}. In \bibinfo{booktitle}{\emph{Proceedings of the 28th
  International Joint Conference on Artificial Intelligence}}.
\newblock


\bibitem[\protect\citeauthoryear{Wu, Wu, Xing, Zhou, and Li}{Wu
  et~al\mbox{.}}{2017}]%
        {wu2017sequential}
\bibfield{author}{\bibinfo{person}{Yu Wu}, \bibinfo{person}{Wei Wu},
  \bibinfo{person}{Chen Xing}, \bibinfo{person}{Ming Zhou}, {and}
  \bibinfo{person}{Zhoujun Li}.} \bibinfo{year}{2017}\natexlab{}.
\newblock \showarticletitle{Sequential Matching Network: A New Architecture for
  Multi-turn Response Selection in Retrieval-Based Chatbots}. In
  \bibinfo{booktitle}{\emph{ACL}}. \bibinfo{pages}{496--505}.
\newblock


\bibitem[\protect\citeauthoryear{Xu, Xu, Zeng, Zhang, Ren, Wang, and Li}{Xu
  et~al\mbox{.}}{2018}]%
        {xu2018unpaired}
\bibfield{author}{\bibinfo{person}{Jingjing Xu}, \bibinfo{person}{SUN Xu},
  \bibinfo{person}{Qi Zeng}, \bibinfo{person}{Xiaodong Zhang},
  \bibinfo{person}{Xuancheng Ren}, \bibinfo{person}{Houfeng Wang}, {and}
  \bibinfo{person}{Wenjie Li}.} \bibinfo{year}{2018}\natexlab{}.
\newblock \showarticletitle{Unpaired Sentiment-to-Sentiment Translation: A
  Cycled Reinforcement Learning Approach}. In
  \bibinfo{booktitle}{\emph{Proceedings of the 56th Annual Meeting of the
  Association for Computational Linguistics (Volume 1: Long Papers)}}.
\newblock


\bibitem[\protect\citeauthoryear{Yan, Song, and Wu}{Yan et~al\mbox{.}}{2016}]%
        {yan2016learning}
\bibfield{author}{\bibinfo{person}{Rui Yan}, \bibinfo{person}{Yiping Song},
  {and} \bibinfo{person}{Hua Wu}.} \bibinfo{year}{2016}\natexlab{}.
\newblock \showarticletitle{Learning to respond with deep neural networks for
  retrieval-based human-computer conversation system}. In
  \bibinfo{booktitle}{\emph{ACM SIGIR}}. \bibinfo{pages}{55--64}.
\newblock


\bibitem[\protect\citeauthoryear{Yang, Hu, Dyer, Xing, and
  Berg-Kirkpatrick}{Yang et~al\mbox{.}}{2018}]%
        {yang2018unsupervised}
\bibfield{author}{\bibinfo{person}{Zichao Yang}, \bibinfo{person}{Zhiting Hu},
  \bibinfo{person}{Chris Dyer}, \bibinfo{person}{Eric~P Xing}, {and}
  \bibinfo{person}{Taylor Berg-Kirkpatrick}.} \bibinfo{year}{2018}\natexlab{}.
\newblock \showarticletitle{Unsupervised text style transfer using language
  models as discriminators}. In \bibinfo{booktitle}{\emph{Advances in Neural
  Information Processing Systems}}.
\newblock


\bibitem[\protect\citeauthoryear{Zeng, Song, Lin, and Sakai}{Zeng
  et~al\mbox{.}}{2019}]%
        {zeng2019attitude}
\bibfield{author}{\bibinfo{person}{Zhaohao Zeng}, \bibinfo{person}{Ruihua
  Song}, \bibinfo{person}{Pingping Lin}, {and} \bibinfo{person}{Tetsuya
  Sakai}.} \bibinfo{year}{2019}\natexlab{}.
\newblock \showarticletitle{Attitude Detection for One-Round Conversation:
  Jointly Extracting Target-Polarity Pairs}. In
  \bibinfo{booktitle}{\emph{Proceedings of the 12th ACM International
  Conference on Web Search and Data Mining}}.
\newblock


\bibitem[\protect\citeauthoryear{Zhang, Dinan, Urbanek, Szlam, Kiela, and
  Weston}{Zhang et~al\mbox{.}}{2018a}]%
        {zhang2018personalizing}
\bibfield{author}{\bibinfo{person}{Saizheng Zhang}, \bibinfo{person}{Emily
  Dinan}, \bibinfo{person}{Jack Urbanek}, \bibinfo{person}{Arthur Szlam},
  \bibinfo{person}{Douwe Kiela}, {and} \bibinfo{person}{Jason Weston}.}
  \bibinfo{year}{2018}\natexlab{a}.
\newblock \showarticletitle{Personalizing Dialogue Agents: I have a dog, do you
  have pets too?}. In \bibinfo{booktitle}{\emph{Proceedings of the 56th Annual
  Meeting of the Association for Computational Linguistics}}.
\newblock


\bibitem[\protect\citeauthoryear{Zhang, Zhu, Wang, Zhao, and Liu}{Zhang
  et~al\mbox{.}}{2019}]%
        {zhang2019neural}
\bibfield{author}{\bibinfo{person}{Wei-Nan Zhang}, \bibinfo{person}{Qingfu
  Zhu}, \bibinfo{person}{Yifa Wang}, \bibinfo{person}{Yanyan Zhao}, {and}
  \bibinfo{person}{Ting Liu}.} \bibinfo{year}{2019}\natexlab{}.
\newblock \showarticletitle{Neural personalized response generation as domain
  adaptation}.
\newblock \bibinfo{journal}{\emph{World Wide Web}} \bibinfo{volume}{22},
  \bibinfo{number}{4} (\bibinfo{year}{2019}).
\newblock


\bibitem[\protect\citeauthoryear{Zhang, Li, Zhu, Zhao, and Liu}{Zhang
  et~al\mbox{.}}{2018b}]%
        {zhang2018modeling}
\bibfield{author}{\bibinfo{person}{Zhuosheng Zhang}, \bibinfo{person}{Jiangtong
  Li}, \bibinfo{person}{Pengfei Zhu}, \bibinfo{person}{Hai Zhao}, {and}
  \bibinfo{person}{Gongshen Liu}.} \bibinfo{year}{2018}\natexlab{b}.
\newblock \showarticletitle{Modeling Multi-turn Conversation with Deep
  Utterance Aggregation}. In \bibinfo{booktitle}{\emph{Proceedings of the 27th
  International Conference on Computational Linguistics}}.
  \bibinfo{pages}{3740--3752}.
\newblock


\bibitem[\protect\citeauthoryear{Zheng, Chen, Zhang, Huang, Mao, and
  Huang}{Zheng et~al\mbox{.}}{2020}]%
        {zheng2020stylized}
\bibfield{author}{\bibinfo{person}{Yinhe Zheng}, \bibinfo{person}{Zikai Chen},
  \bibinfo{person}{Rongsheng Zhang}, \bibinfo{person}{Shilei Huang},
  \bibinfo{person}{Xiaoxi Mao}, {and} \bibinfo{person}{Minlie Huang}.}
  \bibinfo{year}{2020}\natexlab{}.
\newblock \showarticletitle{Stylized dialogue response generation using
  stylized unpaired texts}.
\newblock \bibinfo{journal}{\emph{arXiv preprint arXiv:2009.12719}}
  (\bibinfo{year}{2020}).
\newblock


\bibitem[\protect\citeauthoryear{Zhou, Huang, Zhang, Zhu, and Liu}{Zhou
  et~al\mbox{.}}{2018a}]%
        {DBLP:conf/aaai/ZhouHZZL18}
\bibfield{author}{\bibinfo{person}{Hao Zhou}, \bibinfo{person}{Minlie Huang},
  \bibinfo{person}{Tianyang Zhang}, \bibinfo{person}{Xiaoyan Zhu}, {and}
  \bibinfo{person}{Bing Liu}.} \bibinfo{year}{2018}\natexlab{a}.
\newblock \showarticletitle{Emotional Chatting Machine: Emotional Conversation
  Generation with Internal and External Memory}. In
  \bibinfo{booktitle}{\emph{AAAI-18, New Orleans, Louisiana, USA, February 2-7,
  2018}}. \bibinfo{pages}{730--739}.
\newblock


\bibitem[\protect\citeauthoryear{Zhou, Li, Dong, Liu, Chen, Zhao, Yu, and
  Wu}{Zhou et~al\mbox{.}}{2018b}]%
        {zhou2018multi}
\bibfield{author}{\bibinfo{person}{Xiangyang Zhou}, \bibinfo{person}{Lu Li},
  \bibinfo{person}{Daxiang Dong}, \bibinfo{person}{Yi Liu},
  \bibinfo{person}{Ying Chen}, \bibinfo{person}{Wayne~Xin Zhao},
  \bibinfo{person}{Dianhai Yu}, {and} \bibinfo{person}{Hua Wu}.}
  \bibinfo{year}{2018}\natexlab{b}.
\newblock \showarticletitle{Multi-turn response selection for chatbots with
  deep attention matching network}. In \bibinfo{booktitle}{\emph{ACL}}.
  \bibinfo{pages}{1118--1127}.
\newblock


\bibitem[\protect\citeauthoryear{Zhu, Zhang, Liu, and Wang}{Zhu
  et~al\mbox{.}}{2021}]%
        {zhuneural}
\bibfield{author}{\bibinfo{person}{Qingfu Zhu}, \bibinfo{person}{Weinan Zhang},
  \bibinfo{person}{Ting Liu}, {and} \bibinfo{person}{William~Yang Wang}.}
  \bibinfo{year}{2021}\natexlab{}.
\newblock \showarticletitle{Neural Stylistic Response Generation with
  Disentangled Latent Variables}.
\newblock  (\bibinfo{year}{2021}).
\newblock


\end{thebibliography}

\end{document}